\documentclass[10pt,a4paper]{article}

\usepackage[utf8]{inputenc} 
\usepackage[T1]{fontenc}    
\usepackage{textcomp}       

\usepackage{mathptmx} 
\usepackage{courier}

\usepackage{tipa} 

\newcommand{\safeipa}[1]{\textipa{\upshape#1}}

\DeclareUnicodeCharacter{014B}{\safeipa{\textng}}

\DeclareUnicodeCharacter{0263}{\safeipa{\textgamma}}

\DeclareUnicodeCharacter{03B5}{\safeipa{\textepsilon}}

\DeclareUnicodeCharacter{0283}{\safeipa{\textesh}}

\DeclareUnicodeCharacter{0261}{\safeipa{\textscriptg}}

\DeclareUnicodeCharacter{02A7}{\safeipa{\textteshlig}}

\DeclareUnicodeCharacter{00F8}{\o}

\usepackage[english]{babel}
\usepackage{geometry}
\usepackage{setspace}
\usepackage[affil-it]{authblk}

\usepackage{amsmath}
\usepackage{amsfonts}
\usepackage{seqsplit}
\usepackage{amssymb}

\usepackage{graphicx}
\usepackage{tabularx}
\usepackage{booktabs}
\usepackage{ragged2e}
\usepackage{array}
\usepackage{longtable}
\usepackage{multirow}
\usepackage{tikz-dependency}
\usepackage{subcaption}

\usepackage[hyphens]{url}
\usepackage{natbib}
\renewcommand{\cite}{\citep}
\usepackage{hyperref}
\hypersetup{
	colorlinks=true,
	linkcolor=blue,
	citecolor=blue,
	urlcolor=black
}

\usepackage{fancyhdr}
\title{\textbf{Construction and educational application of a linguistically grounded dependency treebank for Uyghur}}
\author[1,*]{Jiaxin Zuo}
\author[2]{Yiquan Wang}
\author[3]{Yuan Pan}
\author[1,*]{Xiadiya Yibulayin}

\affil[1]{College of Chinese Language and Literature, Xinjiang University, Urumqi, Xinjiang 830046, China}
\affil[2]{College of Mathematics and System Science, Xinjiang University, Urumqi, Xinjiang 830046, China}
\affil[3]{College of Computer Science and Technology, Xinjiang University, Urumqi, Xinjiang 830046, China}
\affil[*]{\textit{Corresponding author:} Jiaxin Zuo (\href{mailto:zuojiaxin@stu.xju.edu.cn}{zuojiaxin@stu.xju.edu.cn}), Xiadiya Yibulayin (\href{mailto:xadiya49@sina.cn}{xadiya49@sina.cn})}
\date{}

\begin{document}
	
	\maketitle
	\thispagestyle{fancy}
	
	\begin{abstract}
		Developing effective educational technologies for low-resource agglutinative languages like Uyghur is often hindered by the mismatch between existing annotation frameworks and specific grammatical structures. To address this challenge, this study introduces the Modern Uyghur Dependency Treebank (MUDT), a linguistically grounded annotation framework specifically designed to capture the agglutinative complexity of Uyghur, including zero copula constructions and fine-grained case marking. Utilizing a hybrid pipeline that combines Large Language Model pre-annotation with rigorous human correction, a high-quality treebank consisting of 3,456 sentences was constructed. Intrinsic structural evaluation reveals that MUDT significantly improves dependency projectivity by reducing the crossing-arc rate from 7.35\% in the Universal Dependencies standard to 0.06\%. Extrinsic parsing experiments using UDPipe and Stanza further demonstrate that models trained on MUDT achieve superior in-domain accuracy and cross-domain generalization compared to UD-based baselines. To validate the practical utility of this computational resource, an AI-assisted grammar tutoring system was developed to translate MUDT-based syntactic analyses into interpretable pedagogical feedback. A controlled experiment involving 35 second-language learners indicated that students receiving syntax-aware feedback achieved significantly higher learning gains compared to those in a control group. These findings establish MUDT as a robust foundation for syntactic analysis and underscore the critical role of linguistically informed natural language processing resources in bridging the gap between computational models and the cognitive needs of second-language learners.
	\end{abstract}
	
	
	\section{Introduction}
	
	Mastering the rich agglutinative morphology of Uyghur presents a formidable challenge for second language (L2) learners. Unlike analytic languages, Uyghur encodes complex grammatical relations---such as mood, voice, and case---within dense suffix chains~\cite{gulinigeer2021morphological}. Learners often struggle with opaque syntactic structures, particularly zero copula constructions where the predicative link is implicit, and the nuanced logical dependencies governing postpositional phrases~\cite{little2001learner}. While Intelligent Computer-Assisted Language Learning (ICALL) systems offer a promising avenue for providing personalized, immediate feedback, their efficacy is strictly bound by the accuracy of the underlying Natural Language Processing (NLP) models~\cite{arnett2025why,settles2020machine}. Unfortunately, for low-resource languages like Uyghur, a significant disconnect remains between the linguistic reality of the language and its computational representation, creating a bottleneck that hinders the development of effective educational technology~\cite{akhundjanova2025parallel,taguchi2022universal,hedderich2021survey}.
	
	Currently, the Universal Dependencies (UD) framework serves as the de facto standard for syntactic annotation due to its cross-linguistic consistency~\cite{demarneffe2021universal,nivre2020universal}. However, this universalist approach often necessitates a trade-off: to maintain typological comparability, it abstracts away language-specific details that are critical for precise grammatical analysis~\cite{osborne2019status}. In the context of Uyghur, the UD framework frequently collapses distinct semantic functions into generic labels and lacks explicit modeling for phenomena such as zero copulas~\cite{taguchi2022universal}. For a learner seeking to understand why a sentence is ungrammatical, a parser based on such simplified standards is insufficient; it leads to structural ambiguities and parsing errors that effectively impose a performance ceiling on downstream applications. Consequently, advancing educational technology for Uyghur requires more than simply accumulating data; it demands a fundamental paradigm shift toward an annotation framework that is grounded in linguistic reality~\cite{mukherjee2006corpus,dash2024corpus}.
	
	To bridge this gap between linguistic theory and educational practice, this study introduces the Modern Uyghur Dependency Treebank (MUDT), a linguistically grounded annotation framework specifically engineered to capture the agglutinative complexity of Uyghur. Diverging from the generic principles of UD, MUDT incorporates a multi-layered morphological decomposition and a tailored dependency taxonomy that explicitly handles zero copulas and fine-grained case marking. We constructed a high-quality treebank of 3,456 sentences through a rigorous hybrid pipeline, combining pre-annotation by Large Language Models with extensive human correction~\cite{gilardi2023chatgpt,alizadeh2023open,bender2021dangers}.
	
	Our evaluation confirms that this language-specific design translates into substantial computational and pedagogical benefits. Intrinsic structural analysis reveals that MUDT significantly simplifies dependency structures, reducing the crossing-arc rate from 7.35\% in the existing UD standard to a negligible 0.06\%, thereby creating a representation that is far more consistent and learnable for algorithms~\cite{futrell2015large,yadav2021do}. Extrinsic parsing experiments using UDPipe and Stanza further demonstrate that models trained on MUDT achieve superior in-domain accuracy and cross-domain generalization. Crucially, to validate the practical utility of this resource, we developed an AI-assisted grammar tutoring system driven by the MUDT parser~\cite{dada2024use}. In a controlled experiment with 35 L2 learners, the system translated syntactic analyses into interpretable pedagogical feedback~\cite{alharbi2023ai}. Results indicated that students receiving this syntax-aware feedback achieved significantly higher learning gains (13.73 points) compared to the control group (7.88 points). These findings establish MUDT not only as a robust foundation for syntactic analysis but also as a critical enabler for the next generation of intelligent language learning tools.
	
	\section{Related Work}
	
	\subsection{Computational Challenges in the Morphology-Syntax Interface}
	The defining characteristic of agglutinative languages like Uyghur is the dense encoding of grammatical categories—such as case, mood, and person—within recursively attached suffixes. Consequently, words in Uyghur function as complete syntactic units that would otherwise require full phrases in analytic languages \cite{tomur2003modern, seddah2013overview, tsarfaty2013parsing}. From a dependency parsing perspective, this shifts the computational burden from linear word order to the internal morphology-syntax interface. However, the lack of explicit boundaries for argument roles, combined with the prevalence of implicit structures like zero copulas, creates systematic ambiguities. Without fine-grained morphosyntactic modeling, parsers struggle to reconstruct predicate-argument structures accurately, often failing to identify the correct head of a phrase \cite{eryigit2008dependency, tusun2022caused}.
	
	These linguistic features introduce dual computational bottlenecks: the infinite productivity of word forms leads to severe data sparsity, while the complex many-to-many mapping between morphology and syntax often traps statistical models in local optima \cite{seddah2013overview, tsarfaty2010statistical}. In the context of ICALL, these parsing deficiencies are particularly detrimental. If a model fails to decode the morphological cues governing argument relations, it cannot distinguish between a learner's error and a valid complex structure. This results in feedback that is either generic or misleading, rather than pedagogically supportive \cite{ragheb2012defining, volodina2014you}. Therefore, explicit and accurate modeling of morphological complexity is a prerequisite for any interpretable educational application \cite{meurers2021natural}.
	
	\subsection{Limitations of Universal Dependencies for Educational Purposes}
	While the Universal Dependencies (UD) framework has successfully facilitated cross-lingual research \cite{nivre2016universal, demarneffe2021universal, croft2017linguistic}, its design philosophy prioritizes typological parallelism over language-specific granularity. To maintain compatibility across diverse languages, the UD standard for Turkic languages often collapses specific case functions into generic labels (e.g., \textit{obl}), effectively obscuring the semantic distinctions carried by functional morphemes \cite{tyers2017assessment, osborne2019status}. 
	
	Furthermore, the lack of standardized guidelines for specific agglutinative phenomena has led to inconsistencies across existing treebanks. Variations in annotating complex constructions, such as post-verbal auxiliaries and pronominalized locatives, hinder the effectiveness of cross-lingual transfer and the development of reliable tools \cite{akhundjanova2025harmonizing, washington2024strategies}. For educational applications, where the goal is to explain why a sentence is constructed in a certain way, this loss of information is critical. A syntax tree that simplifies a complex postpositional phrase into a generic dependent cannot support a tutoring system intended to diagnose learner errors within that very phrase. Thus, while UD provides a structural foundation, a more linguistically grounded annotation scheme is required to meet the precision and interpretability standards of grammar instruction.
	
	\subsection{The Gap in ICALL for Low-Resource Languages}
	Intelligent Computer-Assisted Language Learning (ICALL) systems have demonstrated significant potential in promoting language internalization by focusing learner attention on linguistic forms \cite{amaral2011using, norris2000effectiveness}. However, the distribution of these technologies is structurally unbalanced. Advanced tools capable of deep syntactic analysis are predominantly concentrated in high-resource languages, while low-resource languages face systematic inequalities in both model performance and application coverage \cite{weng2023instructional, blasi2022systematic, pakray2025natural}.
	
	Current applications for under-resourced languages typically rely on shallow processing techniques, such as vocabulary drills or simple pattern matching. While useful for rote memorization, these methods lack the depth to explain systematic syntactic errors or provide corrective feedback on complex sentence structures \cite{bontogon2018intelligent, ward2022how}. From a pedagogical standpoint, effective grammar learning requires ``Focus on Form," where learners can perceive the direct mapping between structure and function. Shallow systems that cannot differentiate argument roles or case functions fail to provide this diagnostic guidance \cite{ragheb2012defining, volodina2014you}. To bridge this gap, educational technologies must be grounded in deep syntactic analysis. Dependency parsing offers an ideal mechanism for visualizing abstract grammatical relations, but only if it is built upon an annotation scheme that accurately reflects the morphosyntactic logic of the target language \cite{meurers2021natural, hellman2019visualizing}.
	
	\section{The MUDT Framework: Bridging Theory and Computation}
	\label{sec:framework}
	
	The primary bottleneck in developing intelligent tutoring systems for Uyghur is not merely the scarcity of data, but the lack of a computational representation that aligns with the linguistic reality faced by learners. Existing universal frameworks, such as Universal Dependencies (UD), often prioritize cross-linguistic consistency over language-specific precision. While valuable for typology, this approach obscures critical grammatical details essential for error diagnosis in an educational setting.
	
	\begin{figure}[htbp]
		\centering
		\includegraphics[width=\textwidth]{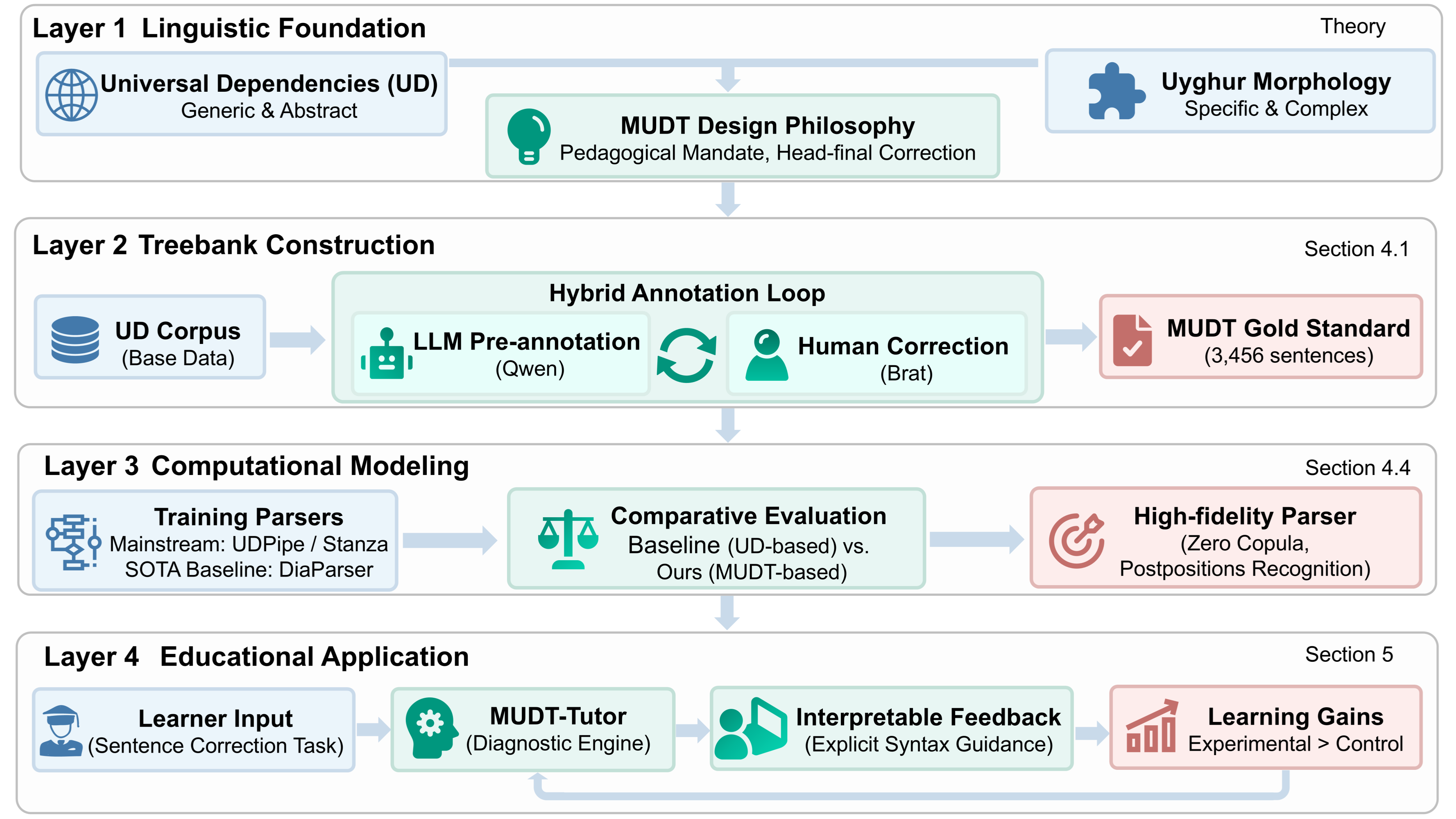}
		\caption{Schematic representation of the full-stack research framework. The framework integrates linguistic theory, resource construction, model training, and educational application through four logical layers. Layer 1 establishes the linguistic foundation by adapting the Universal Dependencies framework to accommodate Uyghur agglutinative features through the MUDT design. Layer 2 details the treebank construction process using a hybrid human-AI loop to produce the gold standard corpus. Layer 3 illustrates the computational modeling phase where various parsers are evaluated to validate the structural fidelity of MUDT. Layer 4 demonstrates the educational application workflow where the system generates interpretable feedback to enhance learner outcomes.}
		\label{fig:framework}
	\end{figure}
	
	To address this, we introduce the Modern Uyghur Dependency Treebank (MUDT) framework, the overall research architecture of which is illustrated in Figure \ref{fig:framework}. Unlike generic approaches, MUDT is designed with a \textit{pedagogical mandate}: the annotation structure must explicitly reflect the morphosyntactic logic that learners are struggling to master.
	
	\subsection{Design Philosophy: Fidelity over Universality}
	Our design philosophy balances computational feasibility with linguistic fidelity~\cite{volodina2014you, gerdes2018sud}. In an educational context, occasional statistical noise in parsing is tolerable, but a \textit{systematic conceptual flaw} in the annotation scheme is fatal. If the annotation scheme incorrectly identifies the head of a phrase (e.g., marking a noun as the head of a postposition), the feedback generated by any downstream AI tool will be pedagogically misleading~\cite{boulton2017corpus}. Therefore, MUDT deviates from UD guidelines specifically where the universal standard fails to capture the agglutinative and head-final nature of Uyghur~\cite{blasi2022over, tyers2017assessment}.
	
	\subsection{Morphological Layer: The Foundation of Syntax}
	Uyghur morphology is the gateway to its syntax~\cite{gulinigeer2021morphological}. A single word form often encodes information that would require a full phrase in English (e.g., \textit{kel-εl-mi-gεn-lik-im-din} means ``because of my inability to come''). 
	
	To support granular feedback, MUDT adopts a four-layer morphological decomposition: (1) Surface Form, (2) Lemma, (3) Morpheme Segmentation, and (4) Morphosyntactic Feature Bundle (FEATS). This layered approach ensures that the parser can ``see'' the internal structure of words, which is a prerequisite for identifying errors in suffix usage---a common stumbling block for learners~\cite{senel2024kardes}.
	
	\subsection{Modeling Key Pedagogical Phenomena}
	\label{sec:key_phenomena}
	The divergence of MUDT from standard UD is most critical in three areas that represent high-frequency error zones for learners: Zero Copulas, Postpositional Phrases, and Compound Predicates. We summarize these structural differences in Table~\ref{tab:comparison_summary} and illustrate the structural contrasts in Figure~\ref{fig:structural_contrast}.
	
	\begin{table}[htb]
		\centering
		\small
		\caption{Contrastive Analysis: How MUDT addresses educational blind spots in UD.}
		\label{tab:comparison_summary}
		\begin{tabularx}{\textwidth}{@{}l X X X@{}}
			\toprule
			\textbf{Phenomenon} & \textbf{UD Approach (The Problem)} & \textbf{MUDT Approach (The Solution)} & \textbf{Educational Benefit} \\ 
			\midrule
			\textbf{Zero Copula} & Treats punctuation or the subject as the root; relation is implicit or lost. & Explicitly marks the nominal predicate as \texttt{root} and the subject link as \texttt{cop:zero}. & Enables system to identify ``Noun Sentences" and provide feedback on missing suffixes. \\ 
			\midrule
			\textbf{Postpositions} & \textit{Noun} $\to$ \textit{Postposition} (\texttt{case}). Noun is the head. & \textit{Postposition} $\to$ \textit{Noun} (\texttt{obj}). Postposition is the head. & Reflects the true governing logic: the postposition determines the case of the noun. \\ 
			\midrule
			\textbf{Compound Predicates} & \textit{Auxiliary} is the \texttt{root}; Lexical verb is a dependent (\texttt{advcl}). & \textit{Lexical Verb} is the \texttt{root}; Auxiliary depends on it (\texttt{aux}). & Focuses analysis on the semantic core (action) rather than grammatical tense markers. \\ 
			\bottomrule
		\end{tabularx}
	\end{table}
	
	\subsubsection{Zero Copula Constructions}
	In Uyghur, the copula is frequently omitted in present tense (e.g., \textit{Bu kitab.} ``This [is] a book"). UD typically struggles here, often attaching the subject to punctuation or leaving the structure fragmented~\cite{bedir2021overcoming} (Figure~\ref{fig:structural_contrast}a). MUDT introduces the \texttt{cop:zero} relation (Figure~\ref{fig:structural_contrast}d). By designating the nominal predicate (\textit{kitab}) as the root, we align the tree with the learner's intuition that ``book" is the core information. This allows an AI tutor to explicitly detect valid verbless sentences.
	
	\subsubsection{Postpositional Logic}
	UD treats postpositions as functional leaves attached to nouns (Figure~\ref{fig:structural_contrast}b). However, for a learner, the postposition \textit{ytʃyn} (for) is the \textit{governor} that demands a specific case from the noun~\cite{osborne2019status}. MUDT inverts this to \texttt{Postposition $\to$ Noun} (Figure~\ref{fig:structural_contrast}e). This structure allows a tutoring system to trace the dependency path upwards and generate feedback like: ``The word \textit{ytʃyn} requires the preceding noun to be in the Nominative case~\cite{johanson2000turkic}."
	
	\subsubsection{Compound Predicates: Reconstructing Argument Structure}
	\label{sec:compound_predicates}
	
	Uyghur frequently employs the ``Converb + Auxiliary" construction to form compound predicates. In the Universal Dependencies standard, the finite auxiliary is regarded as the root, while the lexical verb carrying the core semantics is demoted to an adverbial clause modifier (advcl). The most critical issue is that UD typically attaches the subject directly to the auxiliary, which severs the direct dependency link between the subject and the actual action~\cite{akhundjanova2025parallel}.
	
	MUDT adopts a different strategy. While we still acknowledge the syntactic centrality of the auxiliary as the Root, we mark the lexical verb as a direct complement of the auxiliary rather than a clause. More importantly, we attach core arguments such as the subject and object directly under the lexical verb (Semantic Head) rather than the auxiliary. This ``Auxiliary $\to$ Lexical Verb $\to$ Subject" chain structure successfully distinguishes the Grammatical Head from the Semantic Head~\cite{ivanova2012who}, enabling the pedagogical system to accurately determine who performed what action rather than simply identifying who completed the grammatical aspect.
	
	\begin{figure}[htb]
		\centering
		\small 
		
		% --- Row 1: UD ---
		\begin{subfigure}[b]{0.32\textwidth}
			\centering
			\resizebox{\linewidth}{!}{
				\begin{dependency}[theme = simple, edge unit distance=1.2ex]
					\begin{deptext}[column sep=0.2em]
						Bu \& kitab \& . \\
					\end{deptext}
					\deproot{3}{root}
					\depedge{3}{1}{nsubj}
					\depedge{3}{2}{parataxis}
				\end{dependency}
			}
			\caption{UD: Zero Copula}
			\label{subfig:ud_zero}
		\end{subfigure}
		\hfill
		\begin{subfigure}[b]{0.32\textwidth}
			\centering
			\resizebox{0.8\linewidth}{!}{
				\begin{dependency}[theme = simple, edge unit distance=1.2ex]
					\begin{deptext}[column sep=0.2em]
						Siz \& ytʃyn \\
					\end{deptext}
					\deproot{1}{root} 
					\depedge{1}{2}{case}
				\end{dependency}
			}
			\caption{UD: Noun-Head}
			\label{subfig:ud_post}
		\end{subfigure}
		\hfill
		\begin{subfigure}[b]{0.32\textwidth}
			\centering
			\resizebox{\linewidth}{!}{
				\begin{dependency}[theme = simple, edge unit distance=1.2ex]
					\begin{deptext}[column sep=0.2em]
						Yezip \& boldi \\
					\end{deptext}
					\deproot{2}{root} 
					\depedge{2}{1}{advcl}
				\end{dependency}
			}
			\caption{UD: Aux-Head}
			\label{subfig:ud_compound}
		\end{subfigure}
		
		\vspace{1em} 
		
		% --- Row 2: MUDT ---
		\begin{subfigure}[b]{0.32\textwidth}
			\centering
			\resizebox{\linewidth}{!}{
				\begin{dependency}[theme = simple, edge unit distance=1.2ex]
					\begin{deptext}[column sep=0.2em]
						Bu \& kitab \& . \\
					\end{deptext}
					\deproot{2}{root}
					\depedge{2}{1}{cop:zero}
					\depedge{2}{3}{punct}
				\end{dependency}
			}
			\caption{MUDT: Explicit Pred.}
			\label{subfig:mudt_zero}
		\end{subfigure}
		\hfill
		\begin{subfigure}[b]{0.32\textwidth}
			\centering
			\resizebox{0.8\linewidth}{!}{
				\begin{dependency}[theme = simple, edge unit distance=1.2ex]
					\begin{deptext}[column sep=0.2em]
						Siz \& ytʃyn \\
					\end{deptext}
					\deproot{2}{root} 
					\depedge{2}{1}{obj}
				\end{dependency}
			}
			\caption{MUDT: Post-Head}
			\label{subfig:mudt_post}
		\end{subfigure}
		\hfill
		\begin{subfigure}[b]{0.32\textwidth}
			\centering
			\resizebox{\linewidth}{!}{
				\begin{dependency}[theme = simple, edge unit distance=1.2ex]
					\begin{deptext}[column sep=0.2em]
						Yezip \& boldi \\
					\end{deptext}
					\deproot{2}{root} 
					\depedge{2}{1}{aux}
				\end{dependency}
			}
			\caption{MUDT: Lexical-Head}
			\label{subfig:mudt_compound}
		\end{subfigure}
		
		\caption{Structural comparison of three critical linguistic phenomena. The top row (a-c) shows typical UD structures (often semantically opaque). The bottom row (d-f) shows the proposed MUDT structures designed for pedagogical clarity. Columns correspond to Zero Copula, Postpositional Phrases, and Compound Predicates respectively.}
		\label{fig:structural_contrast}
	\end{figure}
	
	The complete inventory of MUDT dependency relations is provided in Appendix~\ref{app:dependency_inventory}.
	
	\section{Treebank Construction and Technical Validation}
	
	To address the limitations of existing resources and support the development of intelligent educational applications, we constructed the Modern Uyghur Dependency Treebank. This section details the construction pipeline, validates the reliability of the annotation scheme, and empirically demonstrates the superiority of the resulting resource through intrinsic structural metrics, extrinsic parsing experiments, and qualitative error analysis.
	
	\subsection{Data Construction and Annotation Pipeline}
	
	To establish a robust foundation for educational applications while ensuring comparability with previous benchmarks, we utilized the 3,456 sentences from the existing Universal Dependencies UD\_Uyghur-UDT treebank~\cite{eli2016universal}. This corpus represents standard written Uyghur and captures the core grammatical structures essential for language learning~\cite{volodina2014you}. The construction of the treebank followed a hybrid human-AI loop designed to balance computational efficiency with linguistic precision. We initially employed the Qwen3-max Large Language Model via API to generate preliminary annotations based on the MUDT guidelines, but the output exhibited characteristic low-resource processing errors such as hallucinations and inconsistencies in fine-grained case marking~\cite{kellert2025parsing, ding2023gpt3}. Consequently, the pre-annotated data was imported into the Brat annotation platform for a rigorous manual correction phase. A team of six annotators with advanced degrees in linguistics, who underwent specialized training on the MUDT protocols to ensure conceptual alignment, conducted a meticulous line-by-line review to resolve structural ambiguities and correct morphological features. The resulting high-quality corpus was subsequently partitioned into a training set of 1,656 sentences, a development set of 900 sentences, and a test set of 900 sentences for experimental validation.
	
	\subsection{Reliability: Inter-Annotator Agreement}
	
	To ensure that the MUDT annotation scheme is scientifically reproducible and unambiguous, we conducted an inter-annotator agreement study prior to the full-scale annotation~\cite{braylan2022measuring}. We randomly sampled 200 sentences from the corpus, and independent linguists annotated them according to the proposed guidelines.
	
	We evaluated agreement using Unlabeled Attachment Score, Labeled Attachment Score, and Cohen's Kappa coefficient~\cite{zeman2018conll}. The results indicated a high level of consensus, with an Unlabeled Attachment Score of approximately 92.29 percent and a Labeled Attachment Score of 85.45 percent. Furthermore, the Cohen's Kappa coefficient for dependency relations reached 0.9890. These metrics confirm that the proposed annotation scheme is defined with sufficient clarity to allow different human experts to apply it consistently, thereby validating the reliability of the resulting treebank~\cite{artstein2017inter, bayerl2011determines, warrens2015five, mchugh2012interrater}.
	
	\subsection{Intrinsic Evaluation: Structural Quality and Linguistic Plausibility}
	
	Before training computational models, we performed an intrinsic quality assessment to quantify the structural improvements of MUDT compared to the original UD standard~\cite{yadav2021do}. We hypothesize that a linguistically grounded annotation should naturally result in a simpler, more projective graph structure that aligns with the cognitive processing of the language~\cite{gildea2010do,yadav2022assessing}. Using identical dataset splits, we compared the two treebanks across metrics including average dependency distance, crossing arc ratios, and the explicit coverage of linguistic phenomena.
	
	The evaluation results, presented in Table \ref{tab:structural_quality}, indicate that MUDT achieves a fundamental structural optimization. Most notably, the proportion of crossing arcs (non-projective dependencies) decreased drastically from 7.35\% in the UD corpus to a negligible 0.06\% in MUDT. From a linguistic perspective, this reduction is not accidental but stems from resolving the conflict between Uyghur's strict head-final typology and UD's content-head principle. In UD, attaching postpositions or auxiliary verbs as dependents to preceding nouns or verbs often creates ``back-and-forth'' arcs that cross over other modifiers, artificially inflating syntactic complexity. MUDT restores the linear projectivity of the dependency tree by re-assigning functional morphemes (postpositions and light verbs) as syntactic heads, consistent with their roles as case-assigners and mood-markers.
	
	Furthermore, the average dependency distance was reduced from 2.643 to 2.359. In dependency grammar theory, shorter dependency distances correlate with lower cognitive load and higher parsing accuracy~\cite{temperley2007minimization,temperley2018minimizing}. This reduction suggests that the MUDT structure is more local and computationally tractable, theoretically reducing the error propagation in downstream parsing models by keeping related elements closer in the structural hierarchy.
	
	\begin{table}[htb]
		\centering
		\small
		\caption{Comparison of Structural Quality Metrics between UDT and MUDT}
		\label{tab:structural_quality}
		\begin{tabular}{lrrrrr}
			\toprule
			Split & DepDist & Crossing\% & Roots!=1\% & Cycles\% & InvalidHD\% \\
			\midrule
			DEV   & 2.64 $\to$ 2.38 & 8.00 $\to$ 0.11 & 0.00 $\to$ 0.22 & 0.00 $\to$ 0.67 & 0.00 $\to$ 0.56 \\
			TEST  & 2.69 $\to$ 2.37 & 8.67 $\to$ 0.00 & 0.00 $\to$ 0.00 & 0.00 $\to$ 0.00 & 0.00 $\to$ 0.00 \\
			TRAIN & 2.62 $\to$ 2.34 & 6.28 $\to$ 0.06 & 0.00 $\to$ 0.00 & 0.00 $\to$ 0.00 & 0.00 $\to$ 0.00 \\
			\midrule
			TOTAL & 2.64 $\to$ 2.36 & 7.35 $\to$ 0.06 & 0.00 $\to$ 0.06 & 0.00 $\to$ 0.17 & 0.00 $\to$ 0.14 \\
			\bottomrule
		\end{tabular}
	\end{table}
	
	Regarding the representation of specific linguistic phenomena, MUDT provides systematic annotation for zero copula constructions. As shown in Table \ref{tab:feats_comparison}, we identified and labeled 142 instances of the zero copula relation, whereas the UD corpus contained zero instances of this label. This explicit modeling provides necessary supervision signals for educational applications that need to detect valid verbless sentences. Additionally, the cleaning process successfully eliminated structural violations such as multiple roots and cyclic dependencies in the training and test sets, ensuring a normative standard for computational processing.
	
	\begin{table}[htb]
		\centering
		\small
		\caption{Feature Coverage and Phenomenon Explicitation (UDT vs MUDT)}
		\label{tab:feats_comparison}
		\begin{tabular}{lrrr}
			\toprule
			Split & FEATS Coverage \% & FEATS Mean Count & cop:zero Count \\
			\midrule
			DEV   & 44.19 / 44.20 & 0.975 / 0.975 & 0 / 57 \\
			TEST  & 45.02 / 45.02 & 1.032 / 1.032 & 0 / 22 \\
			TRAIN & 46.04 / 46.04 & 1.019 / 1.019 & 0 / 63 \\
			\midrule
			TOTAL & 45.29 / 45.29 & 1.010 / 1.010 & 0 / 142 \\
			\bottomrule
		\end{tabular}
	\end{table}
	
	\subsection{Extrinsic Evaluation: Parser Performance and Generalization}
	
	To empirically verify the computational advantages of the framework, we conducted comparative parsing experiments using three distinct parsers. We selected the transition-based UDPipe and the graph-based Stanza as representative mainstream tools to demonstrate that the performance improvements of MUDT are robust across widely used architectures. This comparison confirms the generalizability of the proposed framework over the Universal Dependencies standard~\cite{qi2020stanza,straka2018udpipe,straka2020udpipe_evalatin}. Furthermore, we employed DiaParser as a state-of-the-art baseline to investigate the performance upper bound~\cite{dozat2017deep,glavas2021supervised}. Utilizing Deep Biaffine Attention, DiaParser achieved a UAS of 0.8050 and a LAS of 0.6675, outperforming the best UDPipe baseline by margins of 7.09\% and 14.05\% respectively. These results suggest that the graph-based Biaffine mechanism is particularly well-suited for modeling the long-distance dependencies inherent in Uyghur's head-final structure, offering improved resolution over transition-based approaches. We subsequently evaluated performance in both in-domain and cross-domain settings to test learnability and generalization.
	
	In the in-domain evaluation, where models were trained and tested on the same scheme, the MUDT-based models consistently outperformed the UD-based baselines. As detailed in Table \ref{tab:parser_performance}, the Stanza model trained on MUDT achieved an Unlabeled Attachment Score of 0.655, significantly higher than the 0.572 achieved by the model trained on UD. This performance gap indicates that the more regular and linguistically grounded structures of MUDT are inherently easier for statistical models to learn.
	
	\begin{table}[htb]
		\centering
		\small
		\caption{Parser Performance Comparison (UAS / LAS)}
		\label{tab:parser_performance}
		\begin{tabular}{llcc}
			\toprule
			Model & Training / Evaluation Setup & UAS & LAS \\
			\midrule
			UDPipe & Train MUDT / Eval MUDT & 0.7341 & 0.5270 \\
			UDPipe & Train UDT / Eval UDT   & 0.7069 & 0.5284 \\
			UDPipe & Train MUDT / Eval UDT  & 0.6084 & 0.2991 \\
			UDPipe & Train UDT / Eval MUDT  & 0.6373 & 0.3252 \\
			\midrule
			Stanza & Train MUDT / Eval MUDT & 0.6554 & 0.2607 \\
			Stanza & Train UDT / Eval UDT   & 0.5722 & 0.2534 \\
			Stanza & Train MUDT / Eval UDT  & 0.5198 & 0.1958 \\
			Stanza & Train UDT / Eval MUDT  & 0.5237 & 0.1585 \\
			\midrule
			\textbf{DiaParser} & \textbf{Train MUDT / Eval MUDT} & \textbf{0.8050} & \textbf{0.6675} \\
			\bottomrule
		\end{tabular}
	\end{table}
	
	We further analyzed the asymmetric compatibility between the two schemes. When the model trained on MUDT was applied to UD data, it maintained a relatively stable structural prediction capability. Conversely, when the model trained on UD was applied to MUDT data, performance degraded sharply, with the Labeled Attachment Score dropping to as low as 0.158 in the Stanza experiment. This asymmetry suggests that MUDT functions as a superset of linguistic knowledge, containing fine-grained syntactic information that the UD model never encountered and therefore cannot predict. This confirms that MUDT captures essential grammatical nuances required for advanced analysis while maintaining backward compatibility with the core structure of the language.
	
	\subsection{Qualitative Analysis of Divergence}
	
	To identify the source of the quantitative performance differences, we analyzed the specific structural divergences between the UD baseline and the MUDT framework. The analysis reveals that the core difference lies in the prioritization of semantic fidelity over formal universality.
	
	Figure \ref{fig:dependency-comparison} illustrates a critical divergence in handling postpositional phrases. The UD-style analysis often inverts the logical relationship to maintain direct arcs, attaching the postposition to the noun. In contrast, MUDT correctly models the postposition as the functional head governing the noun object. This structural clarity allows the MUDT parser to produce logically consistent modification chains, which is essential for error diagnosis in educational settings.
	
	\begin{figure}[htb]
		\centering
		\caption{Structural Comparison. (a) UDT-style analysis where the postposition incorrectly depends on the noun. (b) MUDT analysis correctly modeling the postposition as the head.}
		\label{fig:dependency-comparison}
		
		\begin{subfigure}[b]{\textwidth}
			\centering
			\resizebox{\textwidth}{!}{
				\begin{dependency}[theme = simple]
					\begin{deptext}[column sep=0.8em]
						Nurɣun \& jillar \& ilgiri \& bu \& jεr \& byk-baraqsan \& dεrεxlεr \& bilεn \& qaplanɣan \& quruqluq \& ikεn \\
					\end{deptext}
					\depedge{2}{1}{nummod}
					\depedge{3}{2}{nmod}
					\depedge{11}{3}{advmod}
					\depedge{5}{4}{det}
					\depedge{10}{5}{nmod}
					\depedge{7}{6}{amod}
					\depedge{9}{7}{obl}
					\depedge{7}{8}{case}
					\depedge{10}{9}{amod}
					\depedge{11}{10}{nsubj}
					\deproot{11}{root}
				\end{dependency}
			}
			\caption{UDT-style analysis}
			\label{fig:udt-analysis}
		\end{subfigure}
		
		\vspace{0.5cm} 
		
		\begin{subfigure}[b]{\textwidth}
			\centering
			\resizebox{\textwidth}{!}{
				\begin{dependency}[theme = simple]
					\begin{deptext}[column sep=0.8em]
						Nurɣun \& jillar \& ilgiri \& bu \& jεr \& byk-baraqsan \& dεrεxlεr \& bilεn \& qaplanɣan \& quruqluq \& ikεn \\
					\end{deptext}
					\depedge{2}{1}{nummod}
					\depedge{3}{2}{obj}
					\depedge{5}{4}{det}
					\depedge{10}{5}{nsubj}
					\depedge{7}{6}{amod}
					\depedge{8}{7}{obj}
					\depedge{9}{8}{instr:post}
					\depedge{10}{9}{amod}
					\depedge{10}{3}{post}
					\depedge{11}{10}{cop}
					\deproot{11}{root}
				\end{dependency}%
			}
			\caption{MUDT analysis}
			\label{fig:mudt-analysis}
		\end{subfigure}
	\end{figure}
	
	Further qualitative examination, detailed in Table \ref{tab:qualitative-comparison}, highlights how MUDT explicitly captures phenomena that are often collapsed in the UD standard. For instance, compound predicates are analyzed by identifying the lexical verb as the root rather than the auxiliary, and distinct case relations such as the dative are explicitly labeled rather than being grouped under generic oblique labels. These fine-grained distinctions are what enable the superior performance of the MUDT-trained parsers and provide the necessary granularity for downstream pedagogical feedback.
	
	\begin{table}[htb]
		\centering
		\caption{Qualitative Comparison of Analyses for Key Divergent Phenomena}
		\label{tab:qualitative-comparison}
		\includegraphics[width=\textwidth]{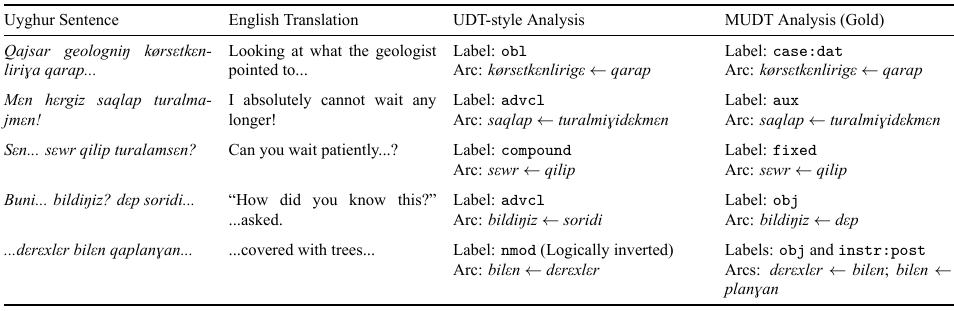}
	\end{table}

	\section{Pedagogical Application: AI-Assisted Grammar Tutoring}
	\label{sec:application}
	
	The ultimate objective of refining the syntactic representation of Uyghur is to facilitate more effective language acquisition. To demonstrate the practical utility of the MUDT framework, we developed an AI-assisted grammar tutoring system and evaluated its effectiveness through a controlled classroom experiment. This section details the system design, the experimental methodology involving L2 learners, and the empirical results regarding learning gains.
	
	\subsection{System Design: From Parsing to Feedback}
	
\begin{figure}[htbp]
		\centering
		\includegraphics[width=\textwidth]{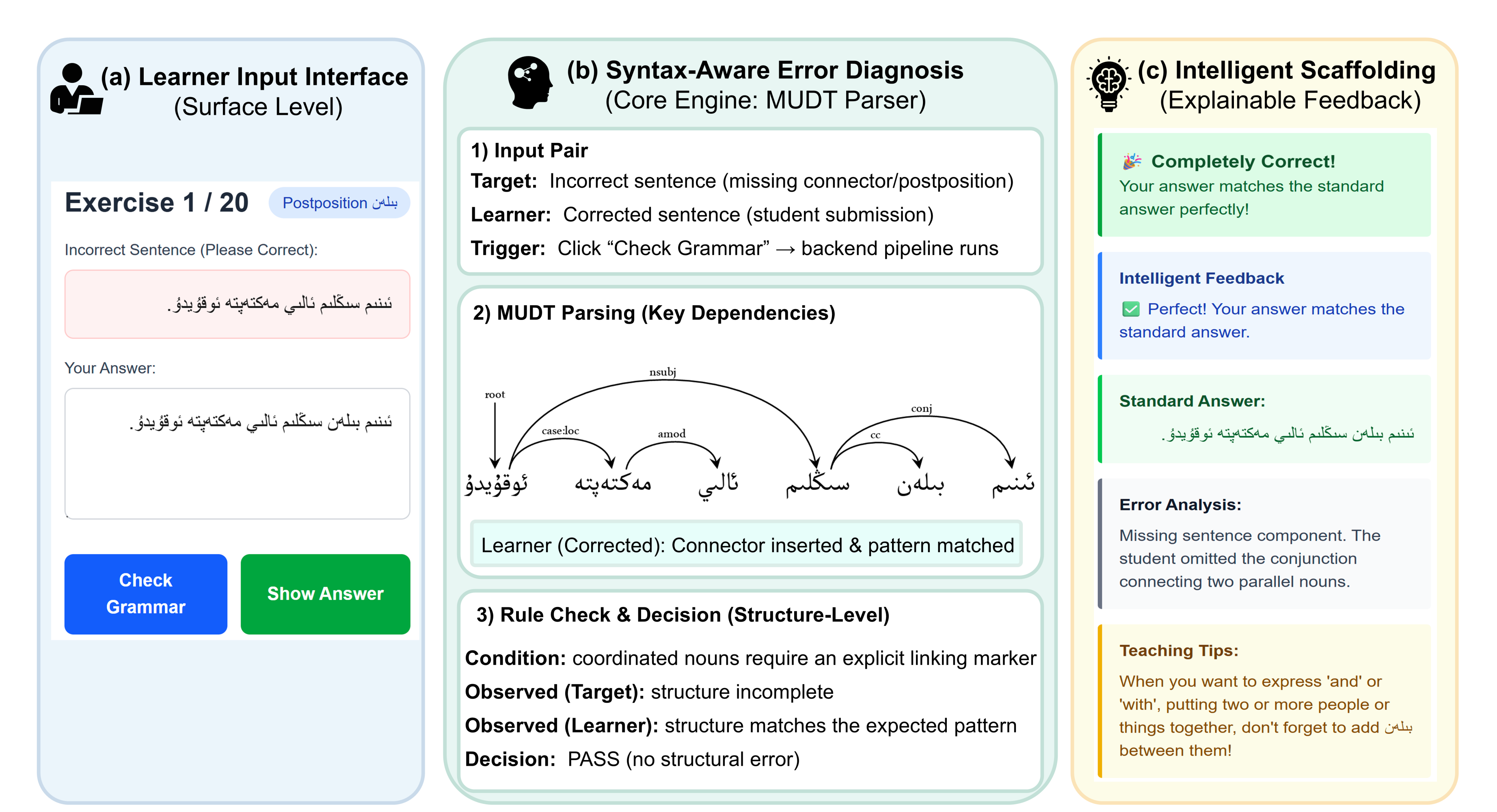}
		\caption{Operational flowchart of the syntactic diagnosis and explainable feedback mechanism in MUDT-Tutor. The process is illustrated using a postposition correction task. (a) The Learner Input Interface presents the front-end interaction where students submit corrections for ungrammatical sentences. (b) The Syntax-Aware Diagnosis module performs parallel parsing of the input pair to extract dependency relations and verify structural validity against the rule engine. (c) The Intelligent Scaffolding Feedback stage generates a dynamic response panel displaying the correctness verdict, standard solution, error classification, and specific pedagogical guidance based on the underlying syntactic analysis.}
		\label{fig:tutor_flow}
	\end{figure}

	The MUDT-Tutor application functions as a comprehensive interactive learning environment rather than a static parser. The operational workflow of the system is depicted in Figure \ref{fig:tutor_flow}. The user interface comprises an exercise display area, a student input field, and an intelligent feedback panel. To evaluate morphosyntactic mastery, the system utilizes a sentence correction format. Learners are presented with a Uyghur sentence containing specific grammatical errors such as incorrect case marking or missing zero copula links. The students then type the corrected version into the input field and initiate the analysis by clicking the submit button.
	
	The submission of an answer triggers a backend pipeline consisting of parsing and diagnosis. The integrated MUDT parser performs real-time syntactic analysis on the input to generate a dependency tree. Subsequently, a rule engine compares this structure against a gold standard tree or a set of predefined error patterns. The primary advantage of this approach lies in the decision authority of the parser. Instead of relying on simple string matching, the system verifies the validity of syntactic relations. For instance, if a student corrects a spelling error but maintains an incorrect dependency between a postposition and its governor, the resulting tree reveals this deep structural flaw. The feedback panel then provides specific structural guidance such as instructions to re-examine the governing object of the postposition instead of showing a generic error message. 
	
	The necessity of the MUDT framework becomes evident when considering specific feedback scenarios. For instance, in zero copula sentences, a standard Universal Dependencies parser typically fails to identify the predicative relationship or attaches the subject to punctuation, rendering the structure opaque to the feedback algorithm. In contrast, the MUDT parser explicitly identifies the \texttt{cop:zero} relation. When the system detects a learner attempting a nominal predicate sentence, it can verify if the subject and predicate are correctly linked and provide positive reinforcement regarding the verbless structure. This allows the system to generate precise hints such as identifying the postposition as the core of the phrase, a level of diagnostic clarity that would be impossible with the generic oblique labels found in previous frameworks.
	
	\subsection{Experimental Design}
	
	To empirically validate the educational impact of the system, we employed a quasi-experimental pre-test and post-test design. The study participants were 35 first-year undergraduate students majoring in Uyghur language at a university in China. All participants were native Chinese speakers who had begun their study of Uyghur three months prior to the experiment, placing them at a novice proficiency level where explicit grammar instruction is most critical. 
	
	The experiment followed a four-day procedure designed to isolate the effect of intelligent feedback. On the first day, all participants completed a paper-based pre-test to establish a baseline of their grammatical competence regarding specific target structures, including case marking and complex predicates. Following the pre-test, participants were randomly assigned to two groups: an experimental group (n=17) and a control group (n=18). On the second day, both groups completed an identical digital exercise consisting of 20 sentence correction tasks. The experimental condition involved the use of the MUDT-Tutor interface, which provided immediate, linguistically grounded feedback upon error submission. This allowed students to engage in a cycle of hypothesis testing and revision. The control condition involved a text-only interface where students completed the same tasks but received no immediate system feedback. To differentiate between short-term memory retention and genuine learning, a cooling-off period was observed on the third day. Finally, on the fourth day, a post-test was administered to measure learning gains. The assessment items on the pre-test and post-test were structurally equivalent but utilized different lexical items to ensure the measurement of grammatical transfer rather than rote memorization~\cite{zhai2021review, alharbi2023ai}.
	
	\subsection{Results and Analysis}
	
	The quantitative analysis focused on the learning gains, defined as the difference between post-test and pre-test scores. To ensure data quality, we performed a data cleaning process excluding participants who exhibited abnormal score regression (post-test score decreasing by more than 5 points compared to pre-test), resulting in a final sample of $N=15$ for the experimental group and $N=16$ for the control group. Descriptive statistics indicated a substantial difference in growth magnitude. The control group improved their average score from 60.75 ($SD=21.97$) to 68.62 ($SD=23.76$), resulting in a mean learning gain of 7.88. In contrast, the experimental group utilizing the MUDT-Tutor improved from a baseline of 57.87 ($SD=20.77$) to a post-test average of 71.60 ($SD=20.24$), achieving a significantly higher mean learning gain of 13.73.
	
	\begin{table}[htb]
		\centering
		\small
		\caption{Comparison of Learning Outcomes between Experimental and Control Groups (Post-cleaning)}
		\label{tab:learning_outcomes}
		\begin{tabular}{lcccccc}
			\toprule
			& & \multicolumn{2}{c}{Pre-test Mean} & \multicolumn{2}{c}{Post-test Mean} & \\
			\cmidrule(lr){3-4} \cmidrule(lr){5-6}
			Group & N & Mean & SD & Mean & SD & Mean Gain \\
			\midrule
			Experimental (MUDT-Tutor) & 15 & 57.87 & 20.77 & 71.60 & 20.24 & 13.73 \\
			Control (No Feedback)     & 16 & 60.75 & 21.97 & 68.62 & 23.76 & 7.88 \\
			\bottomrule
			\multicolumn{7}{l}{\footnotesize \textit{Note:} Significant difference observed: $t(29)=2.51, p=0.018, d=0.90$.}
		\end{tabular}
	\end{table}
	
	The data presented in Table \ref{tab:learning_outcomes} demonstrates that the provision of syntax-aware feedback resulted in superior learning outcomes. To verify the statistical significance, we conducted an independent samples t-test on the learning gains. The analysis revealed a significant difference between the two groups ($t(29) = 2.51, p = 0.018$). Furthermore, the effect size was calculated as Cohen's $d = 0.90$, indicating a large effect. These results confirm that the interpretability of the feedback significantly helped students successfully restructure their interlanguage knowledge compared to the control condition.
	
	Beyond aggregate scores, an analysis of individual ranking changes revealed a positive shift in the relative performance of students in the experimental group. Several students who ranked in the lower quartile during the pre-test exhibited marked improvements in their class ranking during the post-test. For example, students who initially scored below 40 points in the experimental group showed an average improvement that exceeded the class average, suggesting that explicit structural cues are particularly beneficial for struggling learners who may lack the intuition to self-correct. These findings indicate that the linguistic precision of the MUDT framework does not merely serve theoretical completeness but translates directly into measurable pedagogical effectiveness, validating the necessity of linguistically grounded computational resources for low-resource language education.
	
	\section{Discussion}
	
	The results presented in this study highlight the critical link between the quality of computational resources and the effectiveness of educational technology for low-resource languages. Our findings suggest that the misalignment between generic annotation frameworks and the specific agglutinative characteristics of Uyghur constitutes a significant obstacle to developing reliable learning tools. By restructuring the dependency annotation scheme to better align with the internal logic of the language, we observed consistent improvements spanning from intrinsic structural clarity to pedagogical utility.
	
	Intrinsic evaluation of the Modern Uyghur Dependency Treebank suggests that the proposed framework effectively mitigates the structural ambiguities often encountered when applying universal standards to Uyghur syntax. The substantial reduction in crossing arcs indicates that the new scheme aligns more closely with the projectivity principle inherent in the language. This structural consistency appears to facilitate a more straightforward modeling of head-final dependencies. By explicitly annotating the internal structure of compound predicates and postpositional phrases, the framework offers a more transparent hierarchy for these syntactic clusters. Furthermore, the systematic annotation of zero copula constructions provides a clearer representation of nominal sentences, addressing the difficulty previous frameworks faced in assigning valid roots to such structures~\cite{tyers2017assessment, sulubacak2016universal, turk2019improving}.
	
	These structural adjustments appear to contribute to improved computational performance. Parsing experiments indicated that models trained on the linguistically grounded scheme generally outperformed those trained on the universal standard, particularly within complex graph-based architectures. The asymmetry observed in cross-domain tasks suggests that the Modern Uyghur Dependency Treebank may function as a superset of linguistic knowledge, capturing fine-grained morphosyntactic distinctions that are often abstracted in broader universal standards. This implies that preserving these granular details is beneficial for statistical models, as it allows for advanced analysis while retaining the capability to process simpler structures. This observation aligns with prior research demonstrating that preserving fine-grained morphological information enhances syntactic parsing performance, particularly for morphologically rich and agglutinative languages~\cite{seeker2013morphological, park2025enhancing}. Furthermore, studies have shown that the choice of segmentation granularity is critical for optimizing performance across various NLP tasks~\cite{park2025foundations}, suggesting that retaining morphological complexity aids in capturing necessary syntactic dependencies.
	
	The practical relevance of this computational precision was further supported by the pedagogical experiment. The difference in learning gains between the experimental and control groups suggests a correlation between the precision of syntactic analysis and the efficacy of grammar instruction. Students utilizing the intelligent tutoring system demonstrated higher learning outcomes compared to peers who practiced without feedback. This observation indicates that for novice learners, the value of an educational tool is closely tied to the interpretability of its guidance~\cite{heift2022intelligent, son2025artificial, katinskaia2025overview}. While generic error detection provides a useful baseline, its utility may be constrained when it cannot precisely differentiate between functional roles in complex sentences. The ability of our system to provide specific structural diagnoses appeared to help learners test their hypotheses and refine their interlanguage knowledge more effectively.
	
	Beyond technical metrics, this study highlights broader implications for educational equity. Learners of low-resource languages often face challenges accessing the adaptive scaffolding available for dominant languages~\cite{joshi2020state, pakray2025natural, litman2016natural}. By striving to align computational models with the cognitive processing of the learner, specifically by clarifying how functional morphemes govern arguments, the system serves as a supportive cognitive scaffold. This approach may help reduce the difficulty associated with complex agglutinative morphology and encourage novice students to engage with the language more confidently. The outcomes of this study suggest that bridging theoretical linguistics with engineering is a viable pathway for enhancing language education, ensuring that technological tools are better adapted to the specific needs of diverse learner communities.
	
	\section{Conclusion}
	By addressing the structural limitations of universal annotation standards, this study establishes a full-stack methodology that bridges theoretical linguistics with practical educational technology for Uyghur. The construction and evaluation of the Modern Uyghur Dependency Treebank demonstrate that prioritizing linguistic fidelity over typological universality yields a representation that is structurally more coherent and pedagogically transformative. Beyond computational metrics, the true value of this resource lies in its capacity to explicate abstract grammatical relations for the learner. For educators, it provides a reliable auxiliary for error analysis, while for students, it offers a personalized guide through complex morphological structures. These findings confirm that effective intelligent tutoring for low-resource languages relies on computational resources that respect the specific logic of the target language, thereby supporting the preservation and acquisition of the Uyghur language in the digital age.
	
	\section*{Ethics declarations}
	\textbf{Competing interests}\\
	The authors declare no competing interests.\\
	\textbf{Ethical approval}\\
	It will be updated after the article is accepted.\\
	\textbf{Informed consent}\\
	Written informed consent was obtained from all participants prior to data collection, which took place from December 15 to December 18, 2025. All participants were fully informed of the study’s objectives and anonymity measures, and participation was entirely voluntary.
	
	\section*{Data and code availability}
	The Modern Uyghur Dependency Treebank (MUDT) and the source code for the experiment are available at \url{https://github.com/wyqmath/MUDT}.
	
	\section*{Funding}
	This study was supported by the Xinjiang Social Science Foundation Project “Research on the Creative Transformation and Innovative Development of Outstanding Ethnic Minority Cultures in Xinjiang” (Grant No. 2025BYY142). 
	
	\section*{Author Contributions}
	\textbf{Jiaxin Zuo}: Conceptualization, Methodology, Data curation, Formal analysis, Software, Visualization, Writing – original draft, Writing – review \& editing. \\
	\textbf{Yiquan Wang}: Software, Visualization, Writing – original draft, Writing – review \& editing. \\
	\textbf{Yuan Pan}: Software, Writing – original draft, Writing – review \& editing. \\
	\textbf{Xiadiya Yibulayin}: Supervision, Project administration, Resources, Funding acquisition, Writing – review \& editing.
	
	%\bibliographystyle{apalike}
	%\bibliography{references}

	\appendix
	\section{The Complete MUDT Dependency Relation Inventory}
	\label{app:dependency_inventory}
	\begingroup
	\footnotesize
	\setlength{\tabcolsep}{4pt}
	
	\begin{longtable}{@{}p{0.16\textwidth} p{0.18\textwidth} p{0.20\textwidth} p{0.24\textwidth} p{0.20\textwidth}@{}}
		\caption{Dependency Relations in the Proposed Scheme for Modern Uyghur}
		\label{tab:dependency} \\
		\toprule
		\textbf{Dependency Relation} & \textbf{Tag} & \textbf{Example (Uyghur)} & \textbf{Annotation Example} & \textbf{English Translation} \\
		\endfirsthead
		
		\multicolumn{5}{c}%
		{{\bfseries \tablename\ \thetable{} -- continued from previous page}} \\
		\toprule
		\textbf{Dependency Relation} & \textbf{Tag} & \textbf{Example (Uyghur)} & \textbf{Annotation Example} & \textbf{English Translation} \\
		\midrule
		\endhead
		
		\bottomrule
		\multicolumn{5}{r}{{Continued on next page}} \\
		\endfoot
		
		\bottomrule
		\endlastfoot
		
		\midrule
		Adverbial Clause & \texttt{advcl} & U kεlɡεndε, mεn \newline uni køryp qaldim. & \texttt{advcl} kεlɡεndε $\leftarrow$ køryp \newline \texttt{advcl(køryp, kεlɡεndε)} & When he came, I saw him. \\
		Apposition & \texttt{appos} & U oqutquʧi, jεni ustaz. & \texttt{appos} ustaz $\leftarrow$ oqutquʧi \newline \texttt{appos(oqutquʧi, ustaz)} & He is a teacher, that is, a mentor. \\
		Auxiliary & \texttt{aux} & U oqup boldi. & \texttt{aux} oqup $\leftarrow$ boldi \newline \texttt{aux(boldi, oqup)} & He finished reading. \\
		Causal Relation & \texttt{case:abl} & U jεrdin kεldi. & \texttt{case:abl} jεrdin $\leftarrow$ kεldi \newline \texttt{case:abl(kεldi, jεrdin)} & He came from that place. \\
		Locative Relation & \texttt{case:loc} & U jεrdε olturdi. & \texttt{case:loc} jεrdε $\leftarrow$ olturdi \newline \texttt{case:loc(olturdi, jεrdε)} & He sat in that place. \\
		Dative Relation & \texttt{case:dat} & U jεrɡε kεtti. & \texttt{case:dat} jεrɡε $\leftarrow$ kεtti \newline \texttt{case:dat(kεtti, jεrɡε)} & He went to that place. \\
		Possessive Relation & \texttt{case:poss} & Uning kitabi. & \texttt{case:poss} uning $\leftarrow$ kitabi \newline \texttt{case:poss(kitabi, uning)} & His book. \\
		Coordinating Conjunction & \texttt{cc} & U jaki mεn barmaymiz. & \texttt{cc} jaki $\leftarrow$ mεn \newline \texttt{cc(mεn, jaki)} & Either he or I will not go. \\
		Conjunct & \texttt{conj} & Mεn alma, anar, \newline nεʃpyt jedim. & \texttt{conj} alma$\leftarrow$anar$\leftarrow$nεʃpyt \newline \texttt{conj(anar, alma)} \newline \texttt{conj(nεʃpyt, anar)} & I ate an apple, a pomegranate, and a pear. \\
		Copula Relation & \texttt{cop} & U jεxʃi ikεn. & \texttt{cop} jεxʃi $\leftarrow$ ikεn \newline \texttt{cop(ikεn, jεxʃi)} & He is good. \\
		Zero Copula & \texttt{cop:zero} & U oqutquʧi. & \texttt{cop:zero} u $\leftarrow$ oqutquʧi \newline \texttt{cop:zero(oqutquʧi, u)} & He is a teacher. \\
		Determiner & \texttt{det} & Bu kitab. & \texttt{det} bu $\leftarrow$ kitab \newline \texttt{det(kitab, bu)} & This book. \\
		Discourse Element & \texttt{discourse} & Hεε,u, kεldi. & \texttt{discourse} hεε $\leftarrow$ kεldi \newline \texttt{discourse(kεldi, hεε)} & Yes, he came. \\
		Fixed Expression & \texttt{fixed} & Men sizni jaxʃi körimεn. & \texttt{fixed} jaxʃi$\leftarrow$kørimεn \newline \texttt{fixed(kørimεn, jaxʃi)} & I like you. \\
		Instrumental Relation & \texttt{instr:case=loc} & Aptobusda oltur. & \texttt{instr:case=loc} aptobusda $\leftarrow$ oltur \newline \texttt{instr:case=loc(oltur, aptobusda)} & Sit on the bus. \\
		& \texttt{instr:case=dat} & Qolɣa eldi. & \texttt{instr:case=dat} qolɣε $\leftarrow$ eldi \newline \texttt{instr:case=dat(eldi, qolɣε)} & Took it by hand. \\
		& \texttt{instr:case=post} & Qol bilen jεziʃ. & \texttt{instr:case=post} bilen $\leftarrow$ jεziʃ \newline \texttt{instr:case=post(jεziʃ, bilen)} & Write by hand. \\
		Adverbial Modifier & \texttt{advmod} & U tez kεldi. & \texttt{advmod} tez $\leftarrow$ kεldi \newline \texttt{advmod(kεldi, tez)} & He came quickly. \\
		Adjectival Modifier & \texttt{amod} & Jεxʃi kitab. & \texttt{amod} jεxʃi $\leftarrow$ kitab \newline \texttt{amod(kitab, jεxʃi)} & A good book. \\
		Nominal Modifier & \texttt{nmod} & εjnεk istakan. & \texttt{nmod} εjnεk $\leftarrow$ istakan \newline \texttt{nmod(istakan, εjnεk)} & A glass cup. \\
		Nominal Subject & \texttt{nsubj} & Biz kitab oquduq. & \texttt{nsubj} biz $\leftarrow$ oquduq \newline \texttt{nsubj(oquduq, biz)} & We read a book. \\
		Numeric Modifier & \texttt{nummod} & Bir kitab. & \texttt{nummod} bir $\leftarrow$ kitab \newline \texttt{nummod(kitab, bir)} & One book. \\
		Object & \texttt{obj} & Kitabni oquduq. & \texttt{obj} kitabni $\leftarrow$ oquduq \newline \texttt{obj(oquduq, kitabni)} & We read the book. \\
		Postposition & \texttt{post} & U kitab ytʃyn kεldi. & \texttt{post} ytʃyn $\leftarrow$ kεldi \newline \texttt{post(kεldi, ytʃyn)} & He came for the book. \\
		Punctuation & \texttt{punct} & U kεldi. & \texttt{punct} . $\leftarrow$ kεldi \newline \texttt{punct(kεldi, .)} & He came. \\
		Root & \texttt{root} & U kεldi. & \texttt{root} root $\leftarrow$ kεldi \newline \texttt{root(ROOT, kεldi)} & He came. \\
	\end{longtable}
	\endgroup
\end{document}